\documentclass{article} 
\usepackage{iclr2026_conference,times}
\usepackage{pifont}
\DeclareUnicodeCharacter{2717}{\ding{55}}
\usepackage{amssymb}
\usepackage{booktabs}
\usepackage{multirow}
\usepackage{float}   

\usepackage{tikz}
\usepackage{pgfplots}
\pgfplotsset{compat=1.18}
\usepgfplotslibrary{groupplots}
\usepgfplotslibrary{colorbrewer} 
\usepackage{placeins}

\usepackage{float}

\usepackage{amsmath,amsfonts,bm}

\def\eqref#1{equation~\ref{#1}}

\def\1{\bm{1}}

\DeclareMathAlphabet{\mathsfit}{\encodingdefault}{\sfdefault}{m}{sl}
\SetMathAlphabet{\mathsfit}{bold}{\encodingdefault}{\sfdefault}{bx}{n}

\usepackage{hyperref}
\usepackage{url}
\usepackage{graphicx}
\usepackage{tikz}
\usepackage{subcaption}
\usepackage{tcolorbox} 
\usepackage[table,xcdraw]{xcolor}
\usetikzlibrary{decorations.pathmorphing}
\usetikzlibrary{positioning}
\usetikzlibrary{arrows.meta}
\usetikzlibrary{calc}
\usetikzlibrary{fit}
\usetikzlibrary{backgrounds}
\pgfdeclarelayer{background}
\pgfsetlayers{background,main}
\usetikzlibrary{matrix}

\usepackage{tikz}
\usetikzlibrary{positioning,arrows.meta,calc,fit,backgrounds,shapes.misc}

\definecolor{BoxBlue}{RGB}{236,243,255}
\definecolor{BoxYellow}{RGB}{253,248,225}
\definecolor{BoxRed}{RGB}{255,236,236}

\title{CORE-3D: Context-aware Open-vocabulary Retrieval by Embeddings in 3D}

\author{
Mohamad Amin Mirzaei\thanks{These authors contributed equally. For more details, please read the Author Contributions section. }, \
Pantea Amoie\footnotemark[1], \
Ali Ekhterachian\footnotemark[1], \
Matin Mirzababaei\footnotemark[1], \
Babak Khalaj \\ \\ 
Department of Electrical Engineering, Sharif University of Technology \\ \\ 
\texttt{\{mohammad.mirzaii138,pantea.amoie03,} \\ 
\texttt{ali.ekhterachian83,matin.mb,Khalaj\}@sharif.edu}
}

\iclrfinalcopy 

\begin{document}

\maketitle

\begin{abstract}

 Object retrieval from a scene has become a new trend of research due to its numerous applications. Recent approaches achieve zero-shot, open-vocabulary 3D semantic mapping by assigning embedding vectors to 2D class-agnostic masks generated via vision-language models (VLMs) and projecting these into 3D. However, these methods often produce inaccurate semantic assignments due to the direct use of raw masks, limiting their effectiveness in complex environments. To address this, we leverage SemanticSAM with progressive granularity refinement to generate more accurate and numerous object-level masks. To further enhance semantic context, we employ a context-aware CLIP encoding strategy that integrates multiple contextual views of each mask, providing much richer visual context. We also propose a new VLM based framework to process complex 3D queries and retrieve target objects based on 3D position relations between different objects. We evaluate our approach on multiple 3D scene understanding tasks, including 3D semantic segmentation and object retrieval from language queries, across several benchmark datasets. Experimental results demonstrate significant improvements over existing methods, highlighting the effectiveness of our approach.
\end{abstract}

\section{Introduction}

Accurate understanding of 3D environments at the object level is a fundamental requirement for augmented/virtual reality application and etc \citep{anderson2018evaluation, batra2020objectnav, szot2021habitat, gu2024conceptgraphs}.AR/VR systems require precise object-level maps to anchor virtual content in the physical world \citep{lerf2023}. 3D semantic segmentation directly enables these capabilities by assigning category labels to each point in a scene, yielding dense and structured maps that support high-level reasoning and interaction \citep{qi2017pointnet}. Beyond dense labeling, many practical applications require agents not only to segment and recognize objects but also to retrieve specific objects from natural language queries---for example, ``find the chair closest to the table`` or ``locate the vase on the shelf``\citep{chen2020scanrefer, achlioptas2020referit3d}. Such capabilities are essential for interactive agents operating in open and cluttered real-world environments.

\begin{figure}[t]
  \centering
  \includegraphics[width=0.8\linewidth, trim=0 80 0 0, clip]{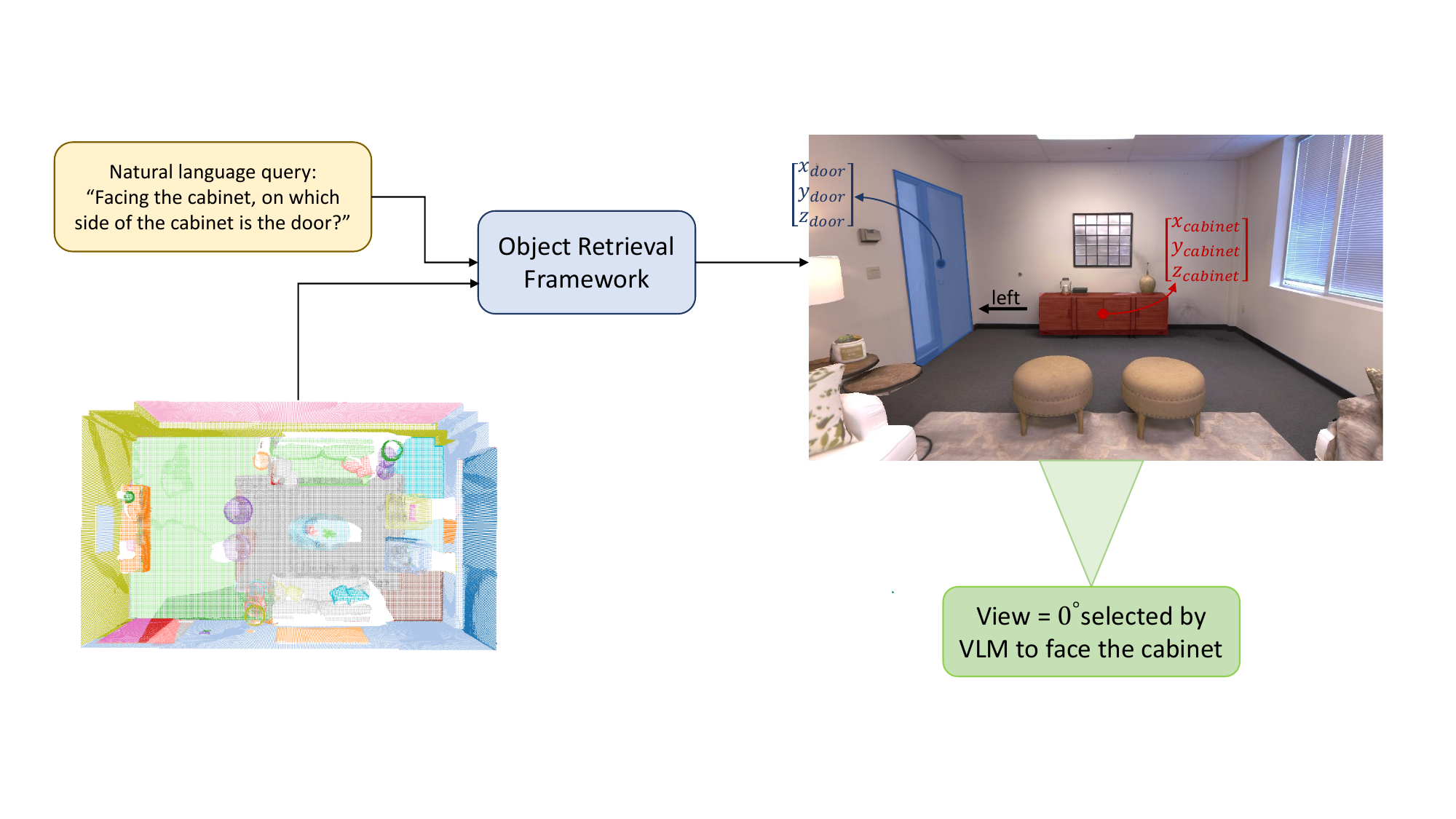}
  \caption{Illustration of object retrieval from natural language in a 3D scene. A natural language query specifies a target and spatial relation (“Facing the cabinet, on which side of the cabinet is the door?”). Our framework retrieves object embeddings, grounds them in 3D coordinates, selects the appropriate view to face the cabinet using VLM, and reasons about spatial orientation to output the correct relation.}
  \label{fig:method}
\end{figure}

Despite progress in supervised 3D scene understanding methods, constructing accurate 3D semantic maps in cluttered, real-world environments remains highly challenging, due to occlusions, incomplete observations, and the prohibitive cost of acquiring large-scale annotated 3D data\citep{patel2025razer,yu2025inst3d}. To reduce reliance on expensive 3D annotations, recent works \citep{gu2024conceptgraphs, jatavallabhula2023conceptfusion} have explored open-vocabulary 3D scene understanding by combining segmentation models with vision–language models. These approaches first extract object masks from 2D images using a segmentation backbone, then assign semantic embeddings to each mask by a vision–language model such as CLIP \citep{radford2021clip}. Projecting these embedded masks into 3D yields semantic maps without requiring task-specific training which enables zero-shot labeling and responding to complex 3D language query tasks. Despite the benefits of these approaches, they encounter some challenges that should be accurately handled. First, 2D segmentation backbones such as SAM \citep{kirillov2023sam} often generate fragmented or incomplete masks, especially in cluttered indoor environments, leading to severe over-segmentation. Second, applying CLIP directly to individual masks provides limited semantic context. Third, aggregating predictions across multiple frames can introduce inconsistencies, as the same object may receive different contextual embeddings depending on viewpoint. As a result, existing foundation-model-based approaches still struggle to construct coherent and reliable 3D semantic maps.

In this work, we present a training-free pipeline that overcomes these challenges by improving both segmentation and embedding generation through progressive refinement and context-aware encoding. First, we leverage SemanticSAM \citep{zhang2023semantic} with progressive granularity adjustment to generate accurate and complete class-agnostic object-level masks, mitigating the fragmentation issues of vanilla SAM. Second, we introduce a context-aware CLIP encoding strategy that aggregates multiple complementary views of each mask with empirically chosen weighting, providing the semantic context necessary for robust classification. Finally, we enforce multi-view consistency by merging overlapping masks in 3D and filtering incomplete or spurious segments with geometric heuristics. Together, these components enable the construction of coherent, high-quality 3D semantic maps in an open-vocabulary, training-free setting, without requiring any 3D supervision.

Beyond 3D semantic segmentation, we extend our framework to object retrieval from natural language instructions. Queries are processed with a large language model (LLM) to extract the target and anchor categories and relational constraints (e.g., ``nearest to the door`` or ``on top of the table``), which are then matched against our fused 3D object embeddings. We also leverage vision large language models to confirm the candidates of target and anchor objects and generate final output by processing the language query together with 3D positions of the confirmed objects. This enables retrieval grounded in both category semantics and spatial relations. 

Our main contributions are:
\begin{itemize}
    \item  We introduce a SemanticSAM refinement strategy that incrementally adjusts granularity, yielding more accurate and complete masks than vanilla SAM.
    \item We propose a context-aware CLIP feature aggregation scheme that combines multiple contextual views of each mask to ensure robust open-vocabulary classification.
    \item We enforce reliable multi-view semantic 3D map by merging overlapping predictions in 3D using geometric and semantic heuristics.
    \item We perform the zero-shot 3D semantic segmentation on Replica \citep{straub2019replica} and ScanNet \citep{dai2017scannet} datasets.
    \item We also extend the pipeline to open-vocabulary 3D object retrieval, using LLM-based query parsing to handle relational constraints in natural language, and evaluate this on the SR3D \citep{achlioptas2020referit3d} benchmark.
    \item Our method achieves superior results on Replica, ScanNet, and SR3D, outperforming prior open-vocabulary approaches in mIoU, f-mIoU, and mAcc for segmentation, while also demonstrating strong retrieval performance. 
\end{itemize}

\section{Related Work}
\label{related_work}

\subsection{Foundation Models for Vision--Language Alignment}
Large-scale vision--language models have become the cornerstone of open-vocabulary perception. 
Contrastive pretraining approaches such as CLIP \citep{radford2021clip} learn aligned image and text embeddings from massive web corpora, enabling zero-shot classification, retrieval, and multimodal reasoning without task-specific finetuning. 
Building on this paradigm, ALIGN \citep{jia2021align} and Florence \citep{yuan2021florence} improved representation quality.  
Subsequently, OVSeg \citep{liang2023ovseg} addresses the CLIP bottleneck in two-stage segmentation by finetuning on masked regions and introducing mask prompt tuning. But even this pipeline can not fully resolve the problem of generating highly enriched embeddings for 2D masks.

\subsection{Foundation Models for Mask Generation}
Complementing semantic embeddings, class-agnostic segmentation priors have shown remarkable generalization across diverse domains. 
MaskFormer \citep{cheng2021maskformer} unified semantic and instance segmentation by reformulating both tasks as per-pixel mask classification, while its successor Mask2Former \citep{cheng2022masked} introduced masked attention to achieve strong panoptic segmentation performance with improved efficiency. 
MaskDINO \citep{li2023maskdino} further integrates detection and segmentation in a unified transformer, showing strong generalization to unseen categories.
The Segment Anything Model (SAM) \citep{kirillov2023sam}, trained on billions of masks, demonstrated that a single backbone can transfer across domains and serves as a universal prior for open-vocabulary pipelines. 
However, SAM often produces fragmented or incomplete masks in cluttered indoor scenes. 
SemanticSAM \citep{zhang2023semantic} alleviates this by generating Segmnet masks with different granularity levels from semantic to instance and to parts.

\subsection{Training-based 3D Scene Understanding}
There are several methods that train a separate network for each scene to predict 3D semantic embeddings or labels for individual 3D points. These approaches typically extract embedding features from 2D image crops or masks of the scene and then assign them to the points. LERF \citep{lerf2023} utilizes the structure of NeRF \citep{mildenhall2020nerf} models to assign not only colors and 3D positions, but also embedding vectors generated by the CLIP model to 3D points. While OpenNeRF \citep{engelmann2024opennerf} leverages OpenSeg \citep{ghiasi2021openseg} feature maps to supervise the 3D feature maps. \newline
Some models blend this idea with 3D Gaussian Splatting \citep{kerbl2023gsplat} structure. LangSplat \citep{qin2024langsplat} uses CLIP embeddings of SAM-generated masks to train an encoder–decoder architecture that produces low-dimensional semantic features for 3D Gaussian distributions. OpenGaussian \citep{wu2024opengaussian}, on the other hand, introduces an intra-mask smoothing loss and an inter-mask contrastive loss to enhance performance on 3D semantic segmentation tasks.

\subsection{Zero-shot 3D Scene Understanding}
Foundation models provide strong 2D priors, and recent works extend open-vocabulary perception into 3D settings, which is crucial for robotics and embodied AI. 
A key trend is integrating vision--language models with 3D representations for mapping, scene understanding, and grounding.
ConceptFusion \citep{jatavallabhula2023conceptfusion} introduces open-set 3D mapping by fusing image features with 3D reconstructions for dense semantic labeling of novel concepts. 
ConceptGraphs \citep{gu2024conceptgraphs} propose open-vocabulary 3D scene graphs that align CLIP features with geometry, supporting perception and planning. 
VoxPoser \citep{huang2023voxposer} applies LLMs and VLMs to synthesize 3D value maps, enabling zero-shot, open-set robot manipulation. 
The Open-Vocabulary Octree-Graph \citep{wang2024octreegraph} uses adaptive octrees to encode occupancy and semantics compactly, while Beyond Bare Queries (BBQ) \citep{linok2025bbq} leverages 3D scene graphs and LLM reasoning for precise language-conditioned object retrieval. 
For robotics, Hierarchical Open-Vocabulary 3D Scene Graphs (HOV-SG) \citep{werby2024hovsg} construct hierarchical floor-room-object graphs for long-horizon language-grounded navigation. 
In 2D, Pixels-to-Graphs (PGSG) \citep{li2024ovsgg} generates scene graphs from images using a generative VLM, supporting both novel relations and downstream vision--language tasks.
These works collectively highlight the importance of combining open-vocabulary semantics with 3D or 2D representations to advance perception, mapping, manipulation, and robot interaction.

\section{Method}
\label{method}

Let $\{(I_t, D_t, \mathbf{T}_t)\}_{t=1}^T$ denote a sequence of RGB images $I_t \in \mathbb{R}^{H \times W \times 3}$, depth maps $D_t \in \mathbb{R}^{H \times W}$, and corresponding camera poses $\mathbf{T}_t \in SE(3)$, where each pose encodes the position and orientation (roll, pitch, yaw) of the camera at time step $t$. Given $D_t$ and $\mathbf{T}_t$, each pixel $p = (u,v)$ can be back-projected into a unique 3D point $\mathbf{x}_p \in \mathbb{R}^3$, thereby allowing us to reconstruct the 3D environment across frames \citep{curless1996volumetric, hartley2003mvgeom, newcombe2011kinectfusion}. Our framework leverages this setup to progressively build semantically meaningful 3D object representations from raw multi-view observations. Starting from 2D instance masks obtained with a refined segmentation strategy, we compute context-aware embeddings that integrate both object-level details and surrounding visual cues. These per-view masks and embeddings are then lifted into 3D, where multi-view consistency and spatial clustering are enforced to merge redundant detections and split over-merged instances. The resulting 3D object candidates are associated with unified embeddings that enable open-vocabulary semantic labeling via CLIP text-image alignment. Finally, we support natural-language object retrieval by grounding query semantics in the labeled 3D scene, allowing objects to be localized based on category, context, and orientation cues.

\begin{figure}[t]
  \centering
  \includegraphics[width=1\linewidth]{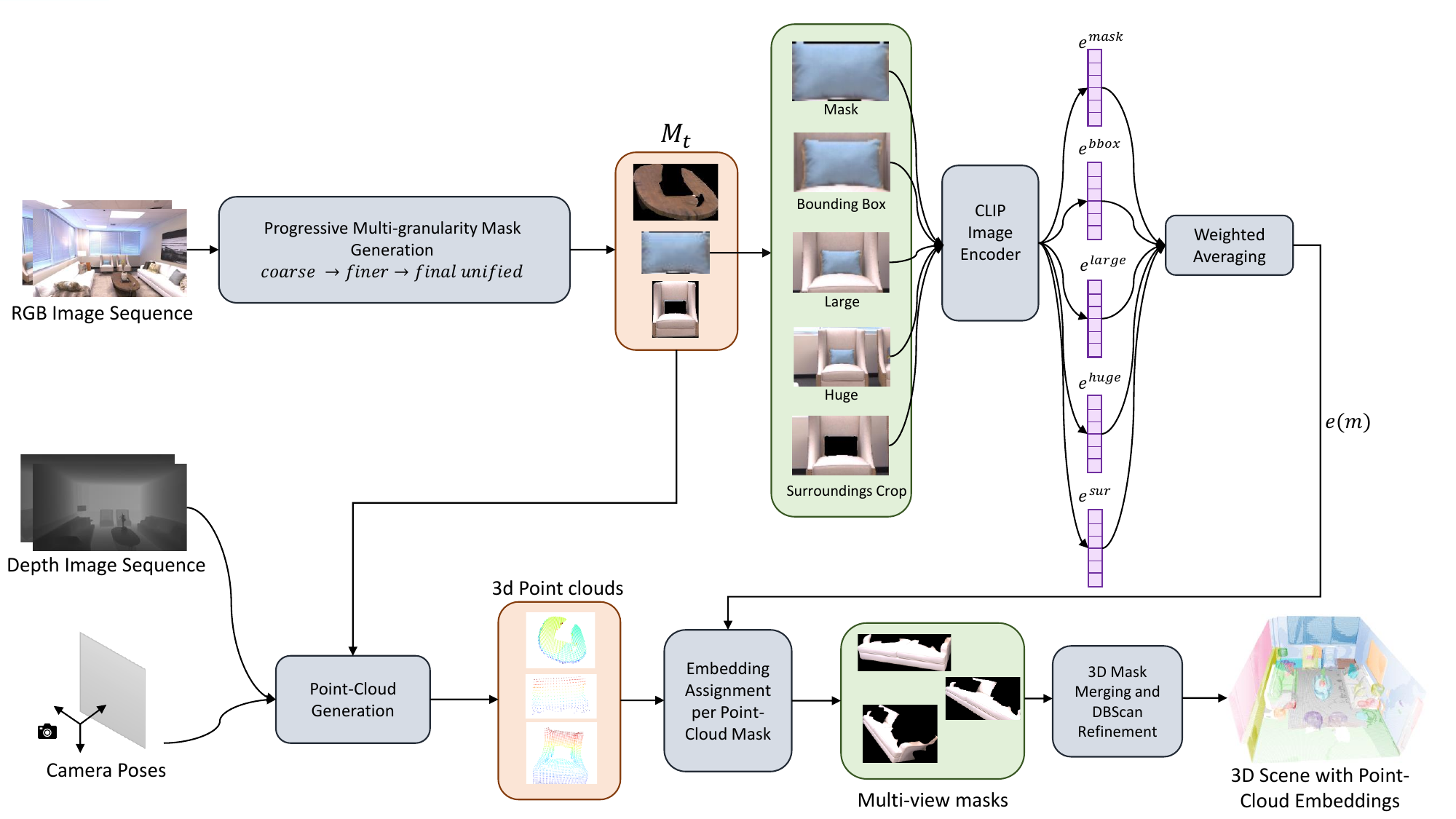}
  \caption{Overview of our training-free open-vocabulary 3D semantic segmentation and retrieval pipeline. Given RGB--D image sequences, we first generate progressive multi-granularity 2D masks ($M_t$) to mitigate fragmentation. Each mask is encoded with CLIP using multiple contextual crops (mask, bounding box, large, huge, surroundings), and their embeddings are aggregated via weighted averaging. In parallel, depth maps and poses are fused into a 3D point cloud, where embeddings are assigned per point-cloud mask. Multi-view predictions are merged and refined with DBSCAN clustering to enforce consistency, resulting in a coherent 3D semantic map with point-cloud embeddings that support both open-vocabulary segmentation and object retrieval.}
  \label{fig:method}
\end{figure}

\subsection{Mask Generation}

Instead of relying on vanilla SAM \citep{kirillov2023sam} for 2D instance segmentation, which often produces \textit{fragmented masks} in cluttered indoor scenes, we employ SemanticSAM \citep{zhang2023semantic}, a variant of SAM that exposes a \textit{granularity parameter} $g \in \mathbb{R}^+$ controlling the scale of segmentation. Small values of $g$ yield coarse masks, while larger values produce finer-grained segments. However, using a single granularity level is suboptimal: coarse values may miss small objects, whereas fine values tend to over-segment large objects into multiple inconsistent parts.

To address this, we introduce a progressive refinement strategy. For each image $I_t$, we generate segmentations at an increasing sequence of granularity levels $\{g_1, g_2, \ldots, g_K\}$. At each step $k$, SemanticSAM produces a set of candidate masks, but we only keep the set of masks $N_k$  with more than a specific threshold of certainty $\tau_{cer}$ denoted as
\[
\mathcal{M}_t^{(k)} = \{m_{t,1}^{(k)}, \ldots, m_{t,N_k}^{(k)}\},
\]
where $N_k$ is the total number of masks generated at granularity level $g_k$.
We then retain only those masks whose area has less than a threshold overlap with any mask discovered at previous levels:
\[
\hat{\mathcal{M}}_t^{(k)} = \Big\{ m \in \mathcal{M}_t^{(k)} \;\big|\; 
\max_{m' \in \cup_{j<k} \hat{\mathcal{M}}_t^{(j)}} 
\frac{|m \cap m'|}{|m|} < \tau_k \Big\},
\]
where $\tau_k \in [0,1]$ denotes the overlap threshold used when adding masks from granularity level $g_k$. 
In practice, $\tau_k$ is varied across levels to balance redundancy removal and coverage: stricter thresholds are applied at coarser levels, while more permissive thresholds are used at finer levels.


This procedure ensures that each new granularity level contributes \textit{novel object candidates} without introducing redundant fragments. Intuitively, large objects are captured at coarse levels, while fine details and small objects are progressively added at higher granularity. By enforcing the threshold $\tau$ and applying these additional filters, we prevent duplicated or noisy masks, leading to a more accurate and complete set of object proposals.

When a single 2D mask spans multiple distinct objects that are spatially close in the 2D image but separated in 3D space (e.g., a vase in front of a couch), we refine it by applying DBSCAN clustering \citep{ester1996dbscan} to each $\hat{\mathcal{M}}_t^{(k)}$ to get the set of $\Tilde{\mathcal{M}}_t^{(k)}$. This step separates the projected 3D points of $m$ into distinct clusters, each corresponding to a potential object. Each cluster is treated as a new candidate instance: we project it back to the image plane, generate the corresponding 2D mask. The final mask set for frame $t$ is then
\[
\mathcal{M}_t = \bigcup_{k=1}^K \Tilde{\mathcal{M}}_t^{(k)}.
\]

\subsection{Context-Aware CLIP Embedding}

Given the refined 2D masks $\mathcal{M}_t$, we next compute semantic embeddings for each object candidate. A direct approach would be to crop the mask region and feed it into CLIP \citep{radford2021clip} or passing a raw 2D mask to OvSeg which is fine tuned on raw masks datasets. However, CLIP relies heavily on visual context, and isolated object crops often lead to ambiguous or incorrect embeddings. However, OvSeg struggles to overcome this issue, but in some scenarios, even an accurate mask does not contain meaningful semantics as shown in Fig.~\ref{fig:method}. To mitigate this, we construct a set of complementary visual crops for each mask that balance object detail with surrounding scene context.

Specifically, for each mask $m$, we extract five complementary crops from the RGB frame $I_t$: 
(i) \textbf{mask crop} ($I^{\text{mask}}$), where pixels outside the mask are set to zero; 
(ii) \textbf{bounding box crop} ($I^{\text{bbox}}$), the tight bounding box enclosing the mask; 
(iii) \textbf{large-context crop} ($I^{\text{large}}$), an expanded bounding box with scale factor $2.5$; 
(iv) \textbf{huge-context crop} ($I^{\text{huge}}$), an expanded bounding box with scale factor $4$; and 
(v) \textbf{surroundings crop} ($I^{\text{sur}}$), obtained by expanding the bounding box with scale factor $3$ and blacking out the mask itself so that only the surrounding environment is visible.

Each of these crops is passed through the CLIP image encoder to obtain embeddings:
\[
\mathbf{e}^{\text{mask}}, \; \mathbf{e}^{\text{bbox}}, \; \mathbf{e}^{\text{large}}, \; \mathbf{e}^{\text{huge}}, \; \mathbf{e}^{\text{sur}} \in \mathbb{R}^d,
\]
where $d$ is the CLIP embedding dimension.

We then compute a context-aware embedding for mask $m$ by taking a weighted combination of these representations:
\[
\mathbf{e}(m) = w_{\text{mask}} \mathbf{e}^{\text{mask}} + w_{\text{bbox}} \mathbf{e}^{\text{bbox}} + w_{\text{large}} \mathbf{e}^{\text{large}} + w_{\text{huge}} \mathbf{e}^{\text{huge}} - w_{\text{sur}} \mathbf{e}^{\text{sur}},
\]
where the weights $\{w_{\text{mask}}, w_{\text{bbox}}, w_{\text{large}}, w_{\text{huge}}, w_{\text{sur}}\}$ are empirically tuned. Note that the surroundings embedding is subtracted with negative weight, enforcing contrastive context by penalizing features dominated by the environment rather than the object itself.

Finally, the embedding is normalized:
\[
\mathbf{e}(m) \leftarrow \frac{\mathbf{e}(m)}{\|\mathbf{e}(m)\|_2}.
\]

These per-view embeddings serve as initial semantic descriptors. During the subsequent 3D merging step, embeddings corresponding to the same physical object observed across multiple views are averaged to form unified object-level representations.

\subsection{3D Mask Merging and Refinement}


Projecting all pixels of mask $m$ yields a 3D point set $\mathcal{X}(m)$, from which we compute its volumetric occupancy $V(m)$ using a voxelization procedure. To consolidate multi-view observations, we evaluate the volumetric intersection between two candidate masks $m_a$ and $m_b$ as
\[
\text{IoV}(m_a,m_b) = \frac{\text{Vol}(\mathcal{X}(m_a) \cap \mathcal{X}(m_b))}{\text{Vol}(\mathcal{X}(m_a))}, \qquad
\text{IoV}(m_b,m_a) = \frac{\text{Vol}(\mathcal{X}(m_a) \cap \mathcal{X}(m_b))}{\text{Vol}(\mathcal{X}(m_b))}.
\]

We merge $m_a$ and $m_b$ into a single 3D object if and only if the following conditions are satisfied:
\[
\text{IoV}(m_a,m_b) > \gamma, \quad \text{IoV}(m_b,m_a) > \gamma, \quad \text{and} \quad 
|\text{IoV}(m_a,m_b) - \text{IoV}(m_b,m_a)| < \delta,
\]
where $\gamma \in [0,1]$ is the minimum overlap threshold and $\delta \in [0,1]$ limits the allowable asymmetry between the two ratios.  

This \textit{symmetric--balanced IoV criterion} ensures that two masks are merged only when they exhibit both high mutual overlap and comparable volumetric support. It prevents degenerate cases where one object is almost fully contained within another but not vice versa---for example, a small cushion lying on a large couch---by rejecting merges with large asymmetry in overlap. 

Along with merging their point clouds, we also average their embeddings:
\[
\mathbf{e}_{\text{merged}} = \frac{1}{n} \sum_{i=1}^{n} \mathbf{e}(m_i),
\]
where $m_1, \ldots, m_n$ denote masks that have been merged.

 and re-apply our multi-crop CLIP embedding procedure. This ensures that each physically distinct object obtains its own semantic descriptor, even if they were originally fused into a single 2D segmentation.

After applying merging, we obtain the final set of refined 3D masks:
\[
\mathcal{M}^{3D} = \{ M_1^{3D}, M_2^{3D}, \ldots, M_{N}^{3D} \},
\]
where each $M_i^{3D}$ is associated with a unified point cloud and an averaged embedding $\mathbf{e}(M_i^{3D})$.

\subsection{Object Retrieval}
\label{sec:object_retrieval}

We extend our pipeline to natural-language object retrieval, where the goal is to localize the specific instance referenced by a free-form query $q$. The task requires reasoning over objects,  their spatial relations, and their orientation cues. Our retrieval pipeline consists of four stages.

\begin{figure}[t]
  \centering
  \includegraphics[width=1\linewidth, height=0.45\linewidth]{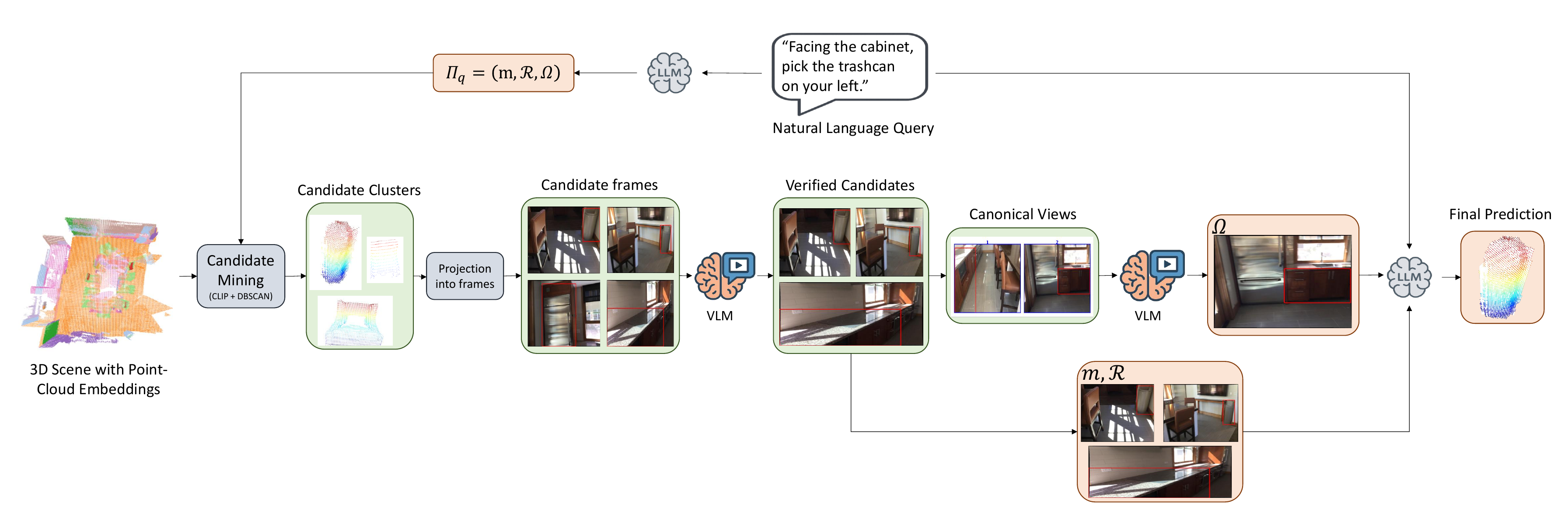}
  \caption{Pipeline for natural-language object retrieval. 
  A free-form query is parsed into structured form $\Pi(q) = (m, \mathcal{R}, \Omega)$. 
  Candidate objects are mined using CLIP similarity and DBSCAN clustering, projected into frames, and verified by a VLM restricted to bounding boxes. 
  If orientation constraints $\Omega$ are present, canonical views are rendered and resolved with a VLM. 
  Finally, an LLM reasons over the verified candidates, referenced objects $\mathcal{R}$, and orientation cues to select the final prediction.}
  \label{fig:ret_method}
\end{figure}

\textbf{Query structuring.}  
We first convert the input query $q$ into a structured form
\[
\Pi(q) = (m,\; \mathcal{R},\; \Omega),
\]
where $m$ denotes the name and attributes of the main object , $\mathcal{R}$ is a set of referenced object names and $\Omega$ encodes orientation constraints (e.g., ``front of the cabinet'').A lightweight LLM extractor produces $\Pi(q)$ deterministically; for the full prompt, see \ref{llm1_prompt}.


\textbf{Candidate mining.}
For each object name $x \in \{m\} \cup \mathcal{R}$, we compute CLIP similarity using pre-computed object embeddings and retain the top-$K$ matches, where $K$ is either a fixed hyperparameter or an upper bound predicted by an LLM based on the object category and scene context. Each match corresponds to a 3D point cluster. We deliberately avoid thresholding CLIP scores and instead always select the top-$K$, since a single similarity threshold that works well for one query can be unreliable across scenes with substantially different similarity distributions. To remove duplicates, we voxelize all candidate clusters and discard any cluster whose voxelized support overlaps a larger neighbor beyond a fixed ratio.


\textbf{View selection and VLM verification.}
Each candidate cluster is then projected into the set of RGB frames. We select the frame that maximizes 3D–2D overlap while penalizing occlusion by other candidate clusters. A VLM is prompted with the bounding box region and asked a binary question to determine whether the object is present. Only candidates passing this check are retained; for the prompt used in this verification step, see ~\ref{vlm_obj_prompt}.


\textbf{Orientation grounding.}
If $\Omega$ specifies an orientation, we collect canonical views of each candidate at discretized yaw bins. These views are tiled into a numbered grid, and a VLM is asked to select the index corresponding to the orientation token (e.g., \texttt{front}). The chosen index is mapped back to a yaw angle and stored with the candidate; for the prompt used in this step, see~\ref{vlm_v_prompt}.

\textbf{Final reasoning.}
The remaining candidates for $m$, together with centroids of related objects and any orientation cues, are passed to an LLM. The LLM receives the original query and structured scene geometry and outputs the index of the final prediction; for the prompt used in this step, see~\ref{llm2_prompt}.

\section{Experiments}
\label{experiments}

We evaluate our framework on two tasks: (i) 3D open-vocabulary semantic segmentation, where the goal is to assign category labels to 3D object instances without task-specific training, and (ii) natural-language object retrieval, where the goal is to localize objects in a 3D scene given free-form text queries that may contain relational and orientation constraints. All experiments are conducted on a workstation equipped with a single NVIDIA RTX4090, with the exception of the VLM and LLM components, which are accessed through external APIs.

\subsection{3D Open-Vocabulary Semantic Segmentation}
\textbf{Datasets.} 
We conduct experiments on two standard benchmarks: Replica \citep{straub2019replica} and ScanNet \citep{dai2017scannet}. 
Replica provides high-quality synthetic RGB-D scans of indoor environments with accurate ground-truth meshes and semantic annotations, while ScanNet consists of large-scale real-world RGB-D sequences with manually annotated 3D semantic and instance labels. 
Following prior work, we use eight Replica scenes: \texttt{room0}, \texttt{room1}, \texttt{room2}, \texttt{office0}, \texttt{office1}, \texttt{office2}, \texttt{office3}, and \texttt{office4}, 
and eight ScanNet scenes: \texttt{0011\_00}, \texttt{0030\_00}, \texttt{0046\_00}, \texttt{0086\_00}, \texttt{0222\_00}, \texttt{0378\_00}, \texttt{0389\_00}, and \texttt{0435\_00}. 
This subset selection ensures comparability with previous zero-shot methods. 
For text–image alignment, we use the Eva02-L CLIP \citep{fang2023eva02, yang2023fgvp} vision-language model.

\textbf{Baselines.}  
We compare against both training-based and zero-shot 3D open-vocabulary segmentation methods. Training-based baselines include LERF \citep{lerf2023}, OpenNeRF \citep{engelmann2024opennerf}, LangSplat \citep{qin2024langsplat}, and OpenGaussian \citep{wu2024opengaussian}.
Zero-shot baselines include ConceptFusion \citep{jatavallabhula2023conceptfusion}, ConceptGraphs \citep{gu2024conceptgraphs}, BBQ-CLIP \citep{linok2025bbq}, OpenMask3D \citep{takmaz2023openmask3d}, and HOV-SG \citep{werby2024hovsg}. 
Note that HOV-SG reports results on a different subset of ScanNet, so we omit its numbers for fair comparison.

\begin{figure*}[t]
  \centering
  \begin{subfigure}[t]{0.16\textwidth}
    \caption*{GT} 
    \includegraphics[width=\linewidth]{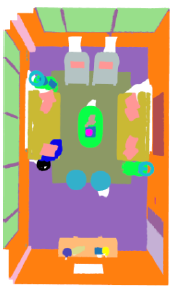}
  \end{subfigure}
  \begin{subfigure}[t]{0.16\textwidth}
    \caption*{Ours}
    \includegraphics[width=\linewidth]{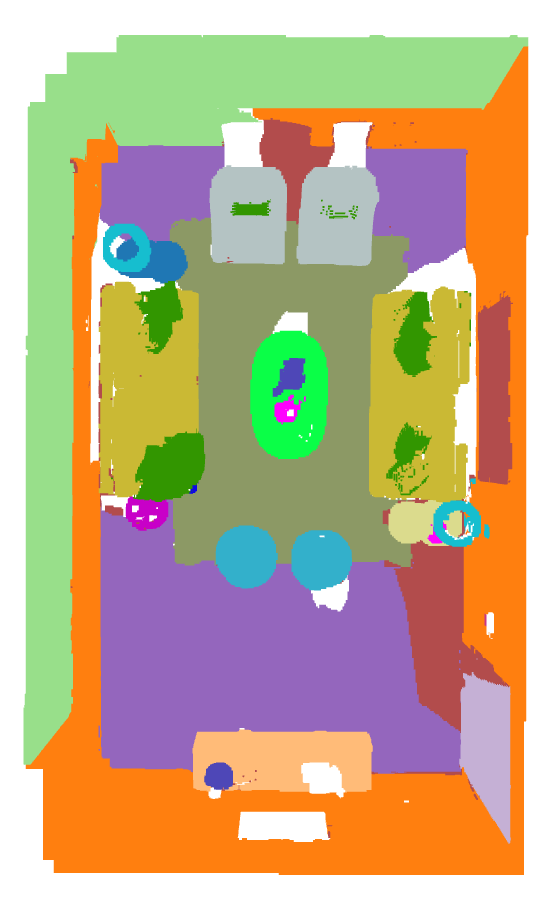}
  \end{subfigure}
  \begin{subfigure}[t]{0.16\textwidth}
    \caption*{BBQ-CLIP}
    \includegraphics[width=\linewidth]{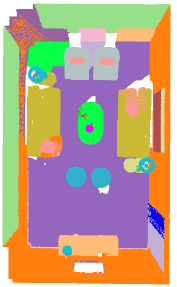}
  \end{subfigure}
  \begin{subfigure}[t]{0.16\textwidth}
    \caption*{ConceptGraphs}
    \includegraphics[width=\linewidth]{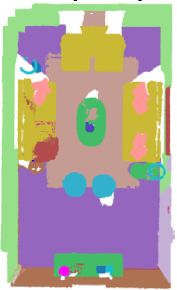}
  \end{subfigure}
  \begin{subfigure}[t]{0.16\textwidth}
    \caption*{ConceptFusion}
    \includegraphics[width=\linewidth]{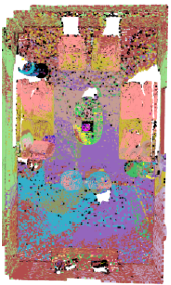}
  \end{subfigure}

    \vspace{0.5em}
    \begin{minipage}{0.95\textwidth}
      \centering
      \setlength{\fboxsep}{0pt} 
      \newcommand{\legenditem}[2]{\textcolor[RGB]{#1}{\rule{1.2ex}{1.2ex}}~#2}
    
      \begin{tabular}{llllllll}
        \legenditem{0,0,0}{basket} & \legenditem{255,187,120}{cabinet} & \legenditem{197,176,213}{door} & \legenditem{140,153,101}{rug} & \legenditem{247,182,210}{pillar} & \legenditem{219,219,141}{plate} & \legenditem{178,76,76}{picture} & \legenditem{196,156,148}{indoor-plant} \\
        \legenditem{91,163,138}{wall-plug} & \legenditem{13,6,230}{blanket} & \legenditem{254,255,0}{candle} & \legenditem{51,176,203}{stool} & \legenditem{255,0,255}{pot} & \legenditem{10,255,71}{table} & \legenditem{148,102,189}{floor} & \legenditem{22,190,207}{lamp} \\
        \legenditem{66,188,102}{plant-stand} & \legenditem{172,114,82}{vent} & \legenditem{152,223,138}{blinds} & \legenditem{180,195,195}{chair} & \legenditem{202,185,52}{sofa} & \legenditem{31,119,180}{book} & \legenditem{200,54,131}{switch} & \legenditem{78,71,183}{vase} \\
        \legenditem{153,98,156}{window} & \legenditem{255,127,15}{wall} & \legenditem{255,151,150}{cushion} & \legenditem{200,0,200}{bin} & \legenditem{50,150,0}{pillow} & & & \\
      \end{tabular}
    \end{minipage}

  \caption{Qualitative comparison of 3D open-vocabulary semantic segmentation on Replica scenes. 
  The GT, BBQ-CLIP, ConceptGraphs, OpenFusion, ConceptFusion columns are adapted from \cite{linok2025bbq}, reproduced here under fair use for research comparison. Our method yields more accurate segmentation boundaries and finer recognition of challenging categories; notably, it is the only method that segments and labels the rug correctly, a class frequently missed or confused by competing approaches.}
  \label{fig:qualitative_replica}
  
\end{figure*}

\textbf{Results.}  
Table~\ref{tab:segmentation_results} reports quantitative comparisons on Replica and ScanNet. Our method achieves the best performance across all evaluated approaches—including both training-based and zero-shot methods. In particular, the improvements in mIoU and fwIoU highlight the benefit of our context-aware embeddings and multi-view 3D refinement, which produce more consistent object representations than 2D-based pipelines. 
We also provide qualitative examples in Fig.~\ref{fig:qualitative_replica}. Visual comparisons on Replica scenes illustrate that our method yields more accurate segmentation boundaries and finer recognition of challenging categories. Our method consistently detects objects across categories with higher fidelity than competing approaches. 

\begin{table}[t]
\caption{3D open-vocabulary semantic segmentation benchmark.}
\label{tab:segmentation_results}
\begin{center}
\begin{tabular}{lcccccc}
\multicolumn{1}{c}{\bf Methods} & \multicolumn{3}{c}{\bf Replica} & \multicolumn{3}{c}{\bf ScanNet} \\
 & mAcc$\uparrow$ & mIoU$\uparrow$ & fmIoU$\uparrow$ & mAcc$\uparrow$ & mIoU$\uparrow$ & fmIoU$\uparrow$ \\
\midrule
\multicolumn{7}{l}{\textit{Training-based methods}} \\
\midrule
LERF & 0.26 & 0.11 & -- & -- & -- & -- \\
OpenNeRF & 0.32 & 0.20 & -- & -- & -- & -- \\
LangSplat & -- & -- & -- & 0.09 & 0.04 & -- \\
OpenGaussian & -- & -- & -- & 0.42 & 0.25 & -- \\
\midrule
\multicolumn{7}{l}{\textit{Zero-shot methods}} \\
\midrule
ConceptFusion & 0.29 & 0.11 & 0.14 & 0.49 & 0.26 & 0.31 \\
OpenMask3D & -- & -- & -- & 0.34 & 0.18 & 0.20 \\
ConceptGraphs & 0.36 & 0.18 & 0.15 & 0.52 & 0.26 & 0.29 \\
HOV-SG & 0.30 & 0.23 & 0.39 & -- & -- & -- \\
BBQ-CLIP & \textbf{0.38} & 0.27 & 0.48 & 0.56 & 0.34 & 0.36 \\
Ours & \textbf{0.38} & \textbf{0.29} & \textbf{0.56} & \textbf{0.61} & \textbf{0.36} & \textbf{0.46} \\
\end{tabular}
\end{center}
\end{table}

\begin{table}[t]
\centering
\caption{\textbf{Grounding accuracy on Sr3D+.}
Accuracy at IoU thresholds A@0.1 and A@0.25 across subsets: Easy, Hard, View-dependent, and View-independent.}
\label{tab:grounding_results}
\setlength{\tabcolsep}{6pt}
\resizebox{\textwidth}{!}{
\begin{tabular}{lcccccccccc}
\multirow{2}{*}{\textbf{Methods}} &
\multicolumn{2}{c}{\textbf{Overall}} &
\multicolumn{2}{c}{\textbf{Easy}} &
\multicolumn{2}{c}{\textbf{Hard}} &
\multicolumn{2}{c}{\textbf{View Dep.}} &
\multicolumn{2}{c}{\textbf{View Indep.}} \\
\cmidrule(lr){2-3}\cmidrule(lr){4-5}\cmidrule(lr){6-7}\cmidrule(lr){8-9}\cmidrule(lr){10-11}
& A@0.1 & A@0.25 & A@0.1 & A@0.25 & A@0.1 & A@0.25 & A@0.1 & A@0.25 & A@0.1 & A@0.25 \\
\midrule
OpenFusion    & 12.6 &  2.4 & 14.0 &  2.4 &  1.3 &  1.3 &  3.8 &  2.5 & 13.7 &  2.4 \\
BBQ-CLIP      & 14.4 &  8.8 & 15.4 &  9.0 &  6.7 &  6.7 & 11.4 &  5.1 & 14.4 &  8.8 \\
ConceptGraphs & 13.3 &  6.2 & 13.0 &  6.8 & 16.0 &  1.3 & 15.2 &  5.1 & 13.1 &  6.4 \\
BBQ           & 34.2 & 22.7 & 34.3 & 22.7 & 33.3 & 22.7 & \textbf{32.9} & 20.3 & 34.4 & 23.0 \\
\textbf{Ours} &   \textbf{41.8}  &   \textbf{35.6}  &   \textbf{41.8}  &   \textbf{35.7}  &   \textbf{41.3}  &   \textbf{34.7}  &   \textbf{32.9}  &   \textbf{30.4}  &   \textbf{43.0}  &   \textbf{36.3}  \\
\end{tabular}
}
\end{table}


\subsection{Natural-Language Object Retrieval}


\textbf{Datasets.} 
We evaluate on the Sr3D+ benchmark \citep{achlioptas2020referit3d}, which provides diverse referring expressions such as relational and orientation-based queries (e.g., ``the table that is far from the armchair'' or ``Facing the cabinet, pick the trashcan on your left.''). 
Following the BBQ \citep{linok2025bbq} setup, we use the same 661 sampled instructions. 
Each query is paired with a ground-truth target (GT) and labeled as \emph{Easy}, \emph{Hard}, \emph{View-dependent}, or \emph{View-independent}, allowing systematic evaluation across reasoning challenges.




\textbf{Baselines.}  
We compare against recent zero-shot 3D grounding approaches, including OpenFusion \citep{yamazaki2024open}, BBQ-CLIP \citep{linok2025bbq}, ConceptGraphs \citep{gu2024conceptgraphs}, and BBQ \citep{linok2025bbq}. 

\textbf{Results.}  
Table~\ref{tab:grounding_results} summarizes quantitative results. 
Our method substantially outperforms all baselines across both IoU thresholds and all difficulty subsets. 
\section{Ablation studies}
In this section, we assess the effectiveness of each part of the 3D semantic segmentation pipeline and also of 3D object retrieval.
First of all, we have evaluated our progressive multi granularity level mask generation strategy. We compare our approach with Vanilla SAM model and also with different single granularity levels of Semantic-SAM model.

\begin{table}[h]
\caption{Granularity-based mask generation impact}
\label{tab:extention_mechanism}
\begin{center}
\begin{tabular}{lcccccc}
\multicolumn{1}{c}{\bf Methods} & \multicolumn{3}{c}{\bf Replica} & \multicolumn{3}{c}{\bf ScanNet} \\
 & mAcc$\uparrow$ & mIoU$\uparrow$ & fmIoU$\uparrow$ & mAcc$\uparrow$ & mIoU$\uparrow$ & fmIoU$\uparrow$ \\
\hline \\[-1.0em]

SAM        & 0.32 & 0.22 & 0.37 & 0.43 & 0.27 & 0.42 \\
Semantic-SAM(Granularity Level=2)        & 0.34 & 0.25 & 0.51 & 0.54 & 0.32 & 0.43 \\
Semantic-SAM(Granularity Level=4)        & \textbf{0.38} & 0.24 & 0.43 & 0.43 & 0.26 & \textbf{0.46} \\
Semantic-SAM(Granularity Level=6)        & 0.26 & 0.11 & 0.14 & 0.17 & 0.09 & 0.16 \\
Ours             & \textbf{0.38} & \textbf{0.29} & \textbf{0.56} & \textbf{0.61} & \textbf{0.36} & \textbf{0.46} \\
\end{tabular}
\end{center}
\end{table}
To evaluate the effectiveness of our context-aware CLIP embedding generation approach, compare it to passing the raw masks to the OvSeg \citep{liang2023ovseg} model which is the finetuned version of the CLIP model on the raw masks dataset.

\begin{table}[h]
\caption{Context-aware CLIP embedding impact}
\label{tab:context-aware-impact}
\begin{center}
\begin{tabular}{lcccccc}
\multicolumn{1}{c}{\bf Methods} & \multicolumn{3}{c}{\bf Replica} & \multicolumn{3}{c}{\bf ScanNet} \\
 & mAcc$\uparrow$ & mIoU$\uparrow$ & fmIoU$\uparrow$ & mAcc$\uparrow$ & mIoU$\uparrow$ & fmIoU$\uparrow$ \\
\hline \\[-1.0em]

OvSeg        & 0.21 & 0.11 & 0.42 & 0.27 & 0.16 & 0.21 \\
Context-Aware CLIP             & \textbf{0.38} & \textbf{0.29} & \textbf{0.56} & \textbf{0.61} & \textbf{0.36} & \textbf{0.46} \\
\end{tabular}
\end{center}
\end{table}
For our approach, we have used Large version of CLIP model which has the same number of parameters as the finetuned OvSeg model. Our approach strongly enhances the CLIP model's embeddings quality without any finetuning.\newline
We have also evaluated our extension mechanism. Instead of relying on target mask size for extending the view around the target mask, we can leverage neighbor masks sizes to determine extension ratio. We define the large extension ratio as the ratio that covers all the neighboring masks and the Huge as the ratio that covers all the neighbors of neighboring masks.

\begin{table}[h]
\caption{Extension mechanism impact.}
\label{tab:extention_mechanism}
\begin{center}
\begin{tabular}{lcccccc}
\multicolumn{1}{c}{\bf Methods} & \multicolumn{3}{c}{\bf Replica} & \multicolumn{3}{c}{\bf ScanNet} \\
 & mAcc$\uparrow$ & mIoU$\uparrow$ & fmIoU$\uparrow$ & mAcc$\uparrow$ & mIoU$\uparrow$ & fmIoU$\uparrow$ \\
\hline \\[-1.0em]

Neighbor coverage        & 0.34 & 0.28 & 0.45 & 0.51 & 0.28 & 0.36 \\
Mask size             & \textbf{0.38} & \textbf{0.29} & \textbf{0.56} & \textbf{0.61} & \textbf{0.36} & \textbf{0.46} \\
\end{tabular}
\end{center}
\end{table}

\section{Conclusion}
\label{sec:conclusion}
We introduced CORE-3D, a training-free pipeline for open-vocabulary 3D perception that combines progressive SemanticSAM refinement, context-aware CLIP embeddings, and multi-view 3D consolidation. This design reduces mask fragmentation, preserves semantic context, and yields coherent object-level maps without requiring 3D supervision. Experiments on Replica and ScanNet show consistent gains in mIoU and fmIoU, while on Sr3D+ our retrieval pipeline—based on structured parsing, VLM verification, and geometric reasoning—achieves clear improvements in grounding accuracy.  
Our results suggest that leveraging richer 2D segmentation and embedding strategies is a powerful alternative to supervision-heavy pipelines, especially in cluttered, open-world environments. Beyond segmentation and retrieval, extending the framework with temporal consistency and deeper integration with multimodal reasoning models could further enhance robustness and generality. In sum, CORE-3D demonstrates that careful refinement and context-rich embeddings make zero-shot 3D mapping and language-grounded retrieval both practical and reliable.


\newpage
\bibliographystyle{iclr2026_conference}
\bibliography{iclr2026_conference}

@article{radford2021clip,
  title     = {Learning Transferable Visual Models From Natural Language Supervision},
  author    = {Alec Radford and Jong Wook Kim and Chris Hallacy and Aditya Ramesh and Gabriel Goh and Sandhini Agarwal and Girish Sastry and Amanda Askell and Pamela Mishkin and Jack Clark and Gretchen Krueger and Ilya Sutskever},
  journal   = {arXiv preprint arXiv:2103.00020},
  year      = {2021},
  url       = {https://arxiv.org/abs/2103.00020}
}

@inproceedings{jia2021align,
  title     = {Scaling Up Visual and Vision-Language Representation Learning With Noisy Text Supervision},
  author    = {Chao Jia and Yinfei Yang and Ye Xia and Yi-Ting Chen and Zarana Parekh and Hieu Pham and Quoc V. Le and Yunhsuan Sung and Zhen Li and Tom Duerig},
  booktitle = {Proceedings of the International Conference on Machine Learning (ICML)},
  year      = {2021}
}

@article{yuan2021florence,
  title     = {Florence: A New Foundation Model for Computer Vision},
  author    = {Lu Yuan and Dongdong Chen and Yi-Ling Chen and Noel Codella and Xiyang Dai and Jianfeng Gao and Houdong Hu and Xuedong Huang and Boxin Li and Chunyuan Li and Ce Liu and Mengchen Liu and Zicheng Liu and Yumao Lu and Yu Shi and Lijuan Wang and Jianfeng Wang and Bin Xiao and Zhen Xiao and Jianwei Yang and Michael Zeng and Luowei Zhou and Pengchuan Zhang},
  journal   = {arXiv preprint arXiv:2111.11432},
  year      = {2021},
  url       = {https://arxiv.org/abs/2111.11432}
}

@inproceedings{kirillov2023sam,
  title     = {Segment Anything},
  author    = {Alexander Kirillov and Eric Mintun and Nikhila Ravi and Hanzi Mao and Chloe Rolland and Laura Gustafson and Tete Xiao and Spencer Whitehead and Alexander C. Berg and Wan-Yen Lo and Piotr Doll{\'a}r and Ross Girshick},
  booktitle = {Proceedings of the IEEE/CVF International Conference on Computer Vision (ICCV)},
  year      = {2023}
}

@article{zhang2023semantic,
  title     = {Semantic-SAM: Segment and Recognize Anything at Any Granularity},
  author    = {Feng Li and Hao Zhang and Peize Sun and Xueyan Zou and Shilong Liu and Jianwei Yang and Chunyuan Li and Lei Zhang and Jianfeng Gao},
  journal   = {arXiv preprint arXiv:2307.04767},
  year      = {2023}
}

@inproceedings{liang2023ovseg,
  title     = {Open-Vocabulary Semantic Segmentation with Mask-adapted CLIP},
  author    = {Feng Liang and Bichen Wu and Xiaoliang Dai and Kunpeng Li and Yinan Zhao and Hang Zhang and Peizhao Zhang and Peter Vajda and Diana Marculescu},
  booktitle = {Proceedings of the IEEE/CVF Conference on Computer Vision and Pattern Recognition (CVPR)},
  year      = {2023},
  doi       = {10.48550/arXiv.2210.04150},
  url       = {https://doi.org/10.48550/arXiv.2210.04150}
}

@inproceedings{li2023maskdino,
  title     = {Mask DINO: Towards a Unified Transformer-based Framework for Object Detection and Segmentation},
  author    = {Feng Li and Hao Zhang and Huaizhe Xu and Shilong Liu and Lei Zhang and Lionel M. Ni and Heung-Yeung Shum},
  booktitle = {Proceedings of the IEEE/CVF Conference on Computer Vision and Pattern Recognition (CVPR)},
  year      = {2023},
  doi       = {10.1109/CVPR52729.2023.00297}, 
  }

@inproceedings{jatavallabhula2023conceptfusion,
  title     = {ConceptFusion: Open-set Multimodal 3D Mapping},
  author    = {Krishna Murthy Jatavallabhula and Alihusein Kuwajerwala and Qiao Gu and Mohd Omama and Tao Chen and Alaa Maalouf and Shuang Li and Ganesh Iyer and Soroush Saryazdi and Nikhil Keetha and Ayush Tewari and Joshua B. Tenenbaum and Celso Miguel de Melo and Madhava Krishna and Liam Paull and Florian Shkurti and Antonio Torralba},
  booktitle = {Robotics: Science and Systems (RSS)},
  year      = {2023},
  doi       = {10.48550/arXiv.2302.07241},
  url       = {https://doi.org/10.48550/arXiv.2302.07241}
}

@inproceedings{gu2024conceptgraphs,
  title     = {ConceptGraphs: Open-Vocabulary 3D Scene Graphs for Perception and Planning},
  author    = {Qiao Gu and Alihusein Kuwajerwala and Sacha Morin and Krishna Murthy Jatavallabhula and Bipasha Sen and Aditya Agarwal and Corban Rivera and William Paul and Kirsty Ellis and Rama Chellappa and Chuang Gan and Celso Miguel de Melo and Joshua B. Tenenbaum and Antonio Torralba and Florian Shkurti and Liam Paull},
  booktitle={2024 IEEE International Conference on Robotics and Automation (ICRA)},
  pages={5021--5028},
  year      = {2024},
  url       = {https://concept-graphs.github.io/}
}

@article{wang2024octreegraph,
  title   = {Open-Vocabulary Octree-Graph for 3D Scene Understanding},
  author  = {Zhigang Wang and Yifei Su and Chenhui Li and Dong Wang and Yan Huang and Bin Zhao and Xuelong Li},
  journal = {arXiv preprint arXiv:2411.16253},   
  year    = {2024},
  url     = {https://arxiv.org/abs/2411.16253}      
}

@article{linok2025bbq,
  title   = {Beyond Bare Queries: Open-Vocabulary Object Grounding with 3D Scene Graph},
  author  = {Sergey Linok and Tatiana Zemskova and Svetlana Ladanova and Roman Titkov and Dmitry Yudin and Maxim Monastyrny and Aleksei Valenkov},
  journal = {arXiv preprint arXiv:2406.07113},
  year    = {2025},
  url     = {https://arxiv.org/abs/2406.07113}
}

@article{werby2024hovsg,
  title   = {Hierarchical Open-Vocabulary 3D Scene Graphs for Language-Grounded Robot Navigation},
  author  = {Abdelrhman Werby and Chenguang Huang and Martin B{\"u}chner and Abhinav Valada and Wolfram Burgard},
  journal = {arXiv preprint arXiv:2403.17846},
  year    = {2024},
  doi     = {10.48550/arXiv.2403.17846},
  note    = {Accepted at RSS 2024}
}

@article{huang2023voxposer,
  title   = {VoxPoser: Composable 3D Value Maps for Robotic Manipulation with Language Models},
  author  = {Wenlong Huang and Chen Wang and Ruohan Zhang and Yunzhu Li and Jiajun Wu and Li Fei-Fei},
  journal = {arXiv preprint arXiv:2307.05973},
  year    = {2023},
  doi     = {10.48550/arXiv.2307.05973}
}

@inproceedings{li2024ovsgg,
  title   = {From Pixels to Graphs: Open-Vocabulary Scene Graph Generation with Vision-Language Models},
  author  = {Rongjie Li and Songyang Zhang and Dahua Lin and Kai Chen and Xuming He},
  booktitle = {Proceedings of the IEEE/CVF Conference on Computer Vision and Pattern Recognition (CVPR)},
  year    = {2024},
  doi     = {10.48550/arXiv.2404.00906}
}

@inproceedings{cheng2021maskformer,
  title     = {Per-Pixel Classification is Not All You Need for Semantic Segmentation},
  author    = {Bowen Cheng and Alexander G. Schwing and Alexander Kirillov},
  booktitle = {Advances in Neural Information Processing Systems (NeurIPS)},
  year      = {2021}
}

@inproceedings{cheng2022masked,
  title     = {Masked-attention Mask Transformer for Universal Image Segmentation},
  author    = {Bowen Cheng and Ishan Misra and Alexander G. Schwing and Alexander Kirillov and Rohit Girdhar},
  booktitle = {Proceedings of the IEEE/CVF Conference on Computer Vision and Pattern Recognition (CVPR)},
  year      = {2022},
  pages     = {1290--1299}
}

@inproceedings{curless1996volumetric,
  title     = {A Volumetric Method for Building Complex Models from Range Images},
  author    = {Brian Curless and Marc Levoy},
  booktitle = {Proceedings of the 23rd Annual Conference on Computer Graphics and Interactive Techniques (SIGGRAPH)},
  year      = {1996},
  pages     = {303--312},
  doi       = {10.1145/237170.237269}
}

@book{hartley2003mvgeom,
  title     = {Multiple View Geometry in Computer Vision},
  author    = {Richard Hartley and Andrew Zisserman},
  publisher = {Cambridge University Press},
  year      = {2003},
  edition   = {2nd}
}

@inproceedings{newcombe2011kinectfusion,
  title     = {KinectFusion: Real-time Dense Surface Mapping and Tracking},
  author    = {Richard A. Newcombe and Shahram Izadi and Otmar Hilliges and David Molyneaux and David Kim and Andrew J. Davison and Pushmeet Kohli and Jamie Shotton and Steve Hodges and Andrew Fitzgibbon},
  booktitle = {Proceedings of the IEEE International Symposium on Mixed and Augmented Reality (ISMAR)},
  year      = {2011},
  pages     = {127--136},
  doi       = {10.1109/ISMAR.2011.6092378}
}

@inproceedings{ester1996dbscan,
  title     = {A Density-Based Algorithm for Discovering Clusters in Large Spatial Databases with Noise},
  author    = {Martin Ester and Hans-Peter Kriegel and J{\"o}rg Sander and Xiaowei Xu},
  booktitle = {Proceedings of the 2nd International Conference on Knowledge Discovery and Data Mining (KDD)},
  year      = {1996},
  pages     = {226--231}
}

@article{straub2019replica,
  title   = {The Replica Dataset: A Digital Replica of Indoor Spaces},
  author  = {Julian Straub and Thomas Whelan and Lingni Ma and Yufan Chen and Erik Wijmans and Simon Green and Jakob J. Engel and Raul Mur-Artal and Carl Ren and Shobhit Verma and Anton Clarkson and Mingfei Yan and Brian Budge and Yajie Yan and Xiaqing Pan and June Yon and Yuyang Zou and Kimberly Leon and Nigel Carter and Jesus Briales and Tyler Gillingham and Elias Mueggler and Luis Pesqueira and Manolis Savva and Dhruv Batra and Hauke M. Strasdat and Renzo De Nardi and Michael Goesele and Steven Lovegrove and Richard Newcombe},
  journal = {arXiv preprint arXiv:1906.05797},
  year    = {2019},
  url     = {https://arxiv.org/abs/1906.05797}
}

@article{dai2017scannet,
  title   = {ScanNet: Richly-annotated 3D Reconstructions of Indoor Scenes},
  author  = {Angela Dai and Angel X. Chang and Manolis Savva and Maciej Halber and Thomas Funkhouser and Matthias Nie{\ss}ner},
  journal = {arXiv preprint arXiv:1702.04405},
  year    = {2017},
  doi     = {10.48550/arXiv.1702.04405},
  url     = {https://arxiv.org/abs/1702.04405}
}

@inproceedings{yang2023fgvp,
  title     = {Fine-Grained Visual Prompting},
  author    = {Lingfeng Yang and Yueze Wang and Xiang Li and Xinlong Wang and Jian Yang},
  booktitle = {Advances in Neural Information Processing Systems (NeurIPS)},
  year      = {2023}
}

@article{fang2023eva02,
  title     = {EVA-02: A Visual Representation for Neon Genesis},
  author    = {Yuxin Fang and Quan Sun and Xinggang Wang and Tiejun Huang and Xinlong Wang and Yue Cao},
  journal   = {arXiv preprint arXiv:2303.11331},
  year      = {2023},
  url       = {https://arxiv.org/abs/2303.11331}
}

@inproceedings{takmaz2023openmask3d,
  title     = {OpenMask3D: Open-Vocabulary 3D Instance Segmentation},
  author    = {Ay{\c{c}}a Takmaz and Elisabetta Fedele and Robert W. Sumner and Marc Pollefeys and Federico Tombari and Francis Engelmann},
  booktitle = {Advances in Neural Information Processing Systems (NeurIPS)},
  year      = {2023},
  journal   = {arXiv preprint arXiv:2306.13631},
  doi       = {10.48550/arXiv.2306.13631},
  url       = {https://doi.org/10.48550/arXiv.2306.13631}
}

@article{anderson2018evaluation,
  title   = {On Evaluation of Embodied Navigation Agents},
  author  = {Peter Anderson and Angel X. Chang and Devendra Singh Chaplot and Alexey Dosovitskiy and Saurabh Gupta and Vladlen Koltun and Jana Kosecka and Jitendra Malik and Roozbeh Mottaghi and Manolis Savva and Amir R. Zamir},
  journal = {arXiv preprint arXiv:1807.06757},
  year    = {2018},
  url     = {https://arxiv.org/abs/1807.06757}
}

@article{batra2020objectnav,
  title   = {ObjectNav Revisited: On Evaluation of Embodied Agents Navigating to Objects},
  author  = {Dhruv Batra and Aaron Gokaslan and Aniruddha Kembhavi and Oleksandr Maksymets and Roozbeh Mottaghi and Manolis Savva and Alexander Toshev and Erik Wijmans},
  journal = {arXiv preprint arXiv:2006.13171},
  year    = {2020},
  url     = {https://arxiv.org/abs/2006.13171}
}

@inproceedings{szot2021habitat,
  title     = {Habitat 2.0: Training Home Assistants to Rearrange their Habitat},
  author    = {Andrew Szot and Alex Clegg and Eric Undersander and Erik Wijmans and Yili Zhao and John Turner and Noah Maestre and Mustafa Mukadam and Devendra Chaplot and Oleksandr Maksymets and Aaron Gokaslan and Vladimir Vondrus and Sameer Dharur and Franziska Meier and Wojciech Galuba and Angel Chang and Zsolt Kira and Vladlen Koltun and Jitendra Malik and Manolis Savva and Dhruv Batra},
  booktitle = {Advances in Neural Information Processing Systems (NeurIPS)},
  year      = {2021}
}

@inproceedings{lerf2023,
  title     = {LERF: Language Embedded Radiance Fields},
  author    = {Justin Kerr and Chung Min Kim and Ken Goldberg and Angjoo Kanazawa and Matthew Tancik},
  booktitle = {Proceedings of the IEEE/CVF International Conference on Computer Vision (ICCV)},
  year      = {2023},
  pages     = {19729--19739},
  month     = oct,
  note      = {Oral Presentation},
  url       = {https://arxiv.org/abs/2303.09553}
}

@inproceedings{qi2017pointnet,
  title     = {PointNet: Deep Learning on Point Sets for 3D Classification and Segmentation},
  author    = {Charles R. Qi and Hao Su and Kaichun Mo and Leonidas J. Guibas},
  booktitle = {Proceedings of the IEEE Conference on Computer Vision and Pattern Recognition (CVPR)},
  year      = {2017},
  pages     = {652--660},
  doi       = {10.1109/CVPR.2017.16}
}

@inproceedings{yamazaki2024open,
  title={Open-fusion: Real-time open-vocabulary 3d mapping and queryable scene representation},
  author={Kashu Yamazaki and Taisei Hanyu and Khoa Vo and Thang Pham and Minh Tran and Gianfranco Doretto and Anh Nguyen and Ngan Le},
  booktitle={2024 IEEE International Conference on Robotics and Automation (ICRA)},
  pages={9411--9417},
  year={2024},
  organization={IEEE}
}

@inproceedings{chen2020scanrefer,
  title     = {ScanRefer: 3D Object Localization in RGB-D Scans using Natural Language},
  author    = {Dave Zhenyu Chen and Angel X. Chang and Matthias Nie{\ss}ner},
  booktitle = {Proceedings of the European Conference on Computer Vision (ECCV)},
  year      = {2020},
  pages     = {2273--2290},
  doi       = {10.1007/978-3-030-58565-5_13},
  url       = {https://doi.org/10.1007/978-3-030-58565-5_13}
}

@inproceedings{achlioptas2020referit3d,
  title     = {ReferIt3D: Neural Listeners for Fine-Grained 3D Object Identification in Real-World Scenes},
  author    = {Panos Achlioptas and Ahmed Abdelreheem and Fei Xia and Mohamed Elhoseiny and Leonidas Guibas},
  booktitle = {Proceedings of the European Conference on Computer Vision (ECCV)},
  year      = {2020},
  pages     = {423--440},
  doi       = {10.1007/978-3-030-58452-8_25},
  url       = {https://doi.org/10.1007/978-3-030-58452-8_25}
}

@article{patel2025razer,
  title={RAZER: Robust Accelerated Zero-Shot 3D Open-Vocabulary Panoptic Reconstruction with Spatio-Temporal Aggregation},
  author={Naman Patel and Prashanth Krishnamurthy and Farshad Khorrami},
  journal={arXiv preprint arXiv:2505.15373},
  year={2025}
}

@inproceedings{yu2025inst3d,
  title={Inst3d-lmm: Instance-aware 3d scene understanding with multi-modal instruction tuning},
  author={Hanxun Yu and Wentong Li and Song Wang and Junbo Chen and Jianke Zhu},
  booktitle={Proceedings of the Computer Vision and Pattern Recognition Conference},
  pages={14147--14157},
  year={2025}
}

@article{mildenhall2020nerf,
  title   = {NeRF: Representing Scenes as Neural Radiance Fields for View Synthesis},
  author  = {Ben Mildenhall and Pratul P. Srinivasan and Matthew Tancik and Jonathan T. Barron and Ravi Ramamoorthi and Ren Ng},
  journal = {arXiv preprint arXiv:2003.08934},
  year    = {2020}
}

@inproceedings{engelmann2024opennerf,
  title     = {OpenNeRF: Open Set 3D Neural Scene Segmentation with Pixel-Wise Features and Rendered Novel Views},
  author    = {Francis Engelmann and Fabian Manhardt and Michael Niemeyer and Keisuke Tateno and Marc Pollefeys and Federico Tombari},
  booktitle = {International Conference on Learning Representations (ICLR)},
  year      = {2024}
}

@inproceedings{qin2024langsplat,
  title     = {LangSplat: 3D Language Gaussian Splatting},
  author    = {Minghan Qin and Wanhua Li and Jiawei Zhou and Haoqian Wang and Hanspeter Pfister},
  booktitle = {Proceedings of the IEEE/CVF Conference on Computer Vision and Pattern Recognition (CVPR)},
  pages     = {20051--20060},
  year      = {2024}
}

@article{wu2024opengaussian,
  title   = {OpenGaussian: Towards Point-Level 3D Gaussian-based Open Vocabulary Understanding},
  author  = {Yanmin Wu and Jiarui Meng and Haijie Li and Chenming Wu and Yahao Shi and Xinhua Cheng and Chen Zhao and Haocheng Feng and Errui Ding and Jingdong Wang and Jian Zhang},
  journal = {arXiv preprint arXiv:2406.02058},
  year    = {2024}
}

@inproceedings{kerbl2023gsplat,
  title     = {3D Gaussian Splatting for Real-Time Radiance Field Rendering},
  author    = {Bernhard Kerbl and Georgios Kopanas and Thomas Leimk{\"u}hler and George Drettakis},
  booktitle = {ACM SIGGRAPH Conference Proceedings},
  year      = {2023}
}

@article{ghiasi2021openseg,
  title   = {Scaling Open-Vocabulary Image Segmentation with Image-Level Labels},
  author  = {Golnaz Ghiasi and Xiuye Gu and Yin Cui and Tsung-Yi Lin},
  journal = {arXiv preprint arXiv:2112.12143},
  year    = {2022}
}


\newpage

\appendix

\section{3D semantic segmentation}
\subsection{Labeling protocol}  
For open-vocabulary labeling, we use the standard CLIP paradigm \citep{radford2021clip}: each refined 3D object mask is encoded into an embedding and compared against a set of text embeddings corresponding to candidate categories. 
The category with the highest similarity is assigned as the predicted label. This procedure enables zero-shot semantic segmentation without task-specific training, while our multi-view refinement makes the assignments more robust to occlusions and clutter.

\subsection{Evaluation protocol}  
We follow the evaluation protocol in prior work. For each predicted point cloud instance, we assign a semantic label by finding the nearest ground-truth instance (based on centroid distance) and transferring its label. 
For each scene, we restrict the text prompts to the classes present in the ground-truth annotations, formatted as “a photo of \texttt{<class name>}.” 
We report mean accuracy (mAcc), mean intersection-over-union (mIoU), and frequency-weighted mIoU (fmIoU).

\section{3D Object Retrieval}

\subsection{Implementation details}
We use the EVA02 CLIP backbone \citep{fang2023eva02} for visual–text alignment, and \texttt{qwen2.5-vl-32b-instruct} as the vision–language model for multimodal encoding. 
Our first LLM, \texttt{gpt-5-mini}, is employed for object extraction, while the second LLM, \texttt{openai-o4-mini}, handles the final decision-making stage.

\subsection{Evaluation protocol}
Following prior work, we report grounding accuracy at two IoU thresholds: A@0.1 and A@0.25. 
A prediction is considered correct if the IoU between the predicted and ground-truth bounding box exceeds the threshold. 
Accuracy is reported overall as well as separately for the four difficulty subsets.

\section{LLM Prompts for Object Retrieval}
\label{sec:llm_prompts}

In this work, we use both LLMs and VLMs to interpret natural-language queries
and convert them into structured outputs for 3D object retrieval. In this
section, we provide the prompts used in our system, covering query
parsing, main-object extraction, and candidate selection.

\subsection{Query Parsing}
\label{llm1_prompt}
The retrieval pipeline begins by converting each natural-language instruction into a structured representation containing the main object, related objects, the spatial relation between them, and any orientation constraints. This is obtained using the following prompt:

\begin{tcolorbox}[colback=gray!5,colframe=gray!40,sharp corners]
Your task is to strictly extract structured information from the given text.

Format rules (follow exactly):
1) Line 1 — MAIN OBJECT: write only one object (the target), optionally followed by comma-separated visual attributes (color, shape, size, texture, material). No non-visual or spatial attributes.

2) Line 2 — RELATED OBJECTS: list all related objects with their visual attributes using:
   object\_name[, attr1, attr2...]; object\_name2[, attr1...]
   Do not repeat the main object. No spatial attributes. Leave blank if none.

3) Line 3 — RELATION KEYWORD: write exactly one of:
   closest, farthest, left, right, front, back, below, supported-by, above, supporting, between

4) Line 4 — ORIENTATION IMPORTANCE: list objects with required orientation in the form:
   object\_name: orientation
   Allowed: front, back, left, right. Leave blank if none.

Do not add placeholders, extra text, labels, punctuation, or formatting.

Text: \texttt{<user\_instruction>}

Return exactly four lines.
\end{tcolorbox}

This prompt provides a deterministic mapping from free-form language to a structured form suitable for geometric reasoning.

Although the prompt requests a spatial-relation keyword on Line 3, this information is not used by our retrieval pipeline. It is included only for analysis, and can be removed without affecting the retrieval system itself.

\subsection{Object Presence Determination}
\label{vlm_obj_prompt}

After selecting the appropriate frame for each candidate, a red bounding box is drawn around its projected 2D region, and the annotated frame is provided to the VLM together with the following prompt:

\begin{tcolorbox}[colback=gray!5,colframe=gray!40,sharp corners]
Focus only on the red bounding box in the image. Ignore everything outside the box. Determine if the object \texttt{<object\_name>} is present inside the box. It is valid even if only part of the object appears, as long as the main subject is included. Answer strictly with 'Yes' or 'No'.
\end{tcolorbox}

This prompt ensures that the VLM evaluates the candidate purely based on the content of the region indicated by the bounding box, while visually ignoring the rest of the frame.

\subsection{Viewpoint Identification}
\label{vlm_v_prompt}



For instructions that include orientation-dependent relations (such as “front of,” “left of,” or “to the right when facing the cabinet”), the system must determine which viewpoint of the candidate object corresponds to the requested orientation. To do this, we render viewpoints at sufficiently separated yaw angles (e.g., 90° increments), determined by camera locations relative to the object. These views are arranged into a single grid image, with each sub-image labeled by an index and containing a red bounding box marking the target object.

This grid image is then given to the VLM, which selects the sub-image matching the orientation specified in the instruction. The prompt used is:

\begin{tcolorbox}[colback=gray!5,colframe=gray!40,sharp corners]

In the provided image, multiple sub-images are arranged in a grid, each bordered in blue and labeled with a number at the top. In every sub-image, the target object '\texttt{<object\_name>}' is indicated with a red bounding box. Ignore all other content. Determine which sub-image shows the object from the '\texttt{<orientation>}' view. Output only the index number above that sub-image.
\end{tcolorbox}

This allows the VLM to choose the correct viewpoint among all candidates presented within the same image grid.

\subsection{Candidate Selection and Final Retrieval}
\label{llm2_prompt}

Once presence verification and viewpoint identification  (when required) are completed, the system must choose the correct 3D instance of the main object. To do so, an LLM is provided with the original instruction together with the 3D centroids of all candidate object clusters. Each centroid corresponds to the geometric center of the candidate’s projected 3D point cloud. The LLM receives these centroids in plain text form along with the positions of referenced objects and, when applicable, the estimated orientation angles.

Depending on whether the instruction requires orientation reasoning, one of the two prompts below is used.

\paragraph{Without orientation.}
If the instruction does not involve orientation-dependent relations, the LLM receives the following prompt:

\begin{tcolorbox}[colback=gray!5,colframe=gray!40,sharp corners]
Text: \texttt{<user\_instruction>}

The main object is \texttt{<main>}.
It has the following candidate positions with indices (x, y, z):

Index 0: (x=\texttt{<x0>}, y=\texttt{<y0>}, z=\texttt{<z0>})

Index 1: (x=\texttt{<x1>}, y=\texttt{<y1>}, z=\texttt{<z1>})

Index 2: (x=\texttt{<x2>}, y=\texttt{<y2>}, z=\texttt{<z2>})

\ldots

Other objects and their positions are:

\texttt{<obj>}[\texttt{<k>}]: (x=\texttt{<x>}, y=\texttt{<y>}, z=\texttt{<z>})

\ldots

Instruction:
Based on the text above, decide which candidate index of \texttt{<main>} is the correct target.
Return only the numeric index (0, 1, 2, \ldots). Do not return anything else.
\end{tcolorbox}

Here, all coordinate placeholders (e.g., \texttt{<x0>}, \texttt{<y>}) are filled with the centroids of the 3D clusters, and name placeholders (e.g., \texttt{<main>}, \texttt{<obj>}) are filled with the corresponding object identifiers.

\paragraph{With orientation.}
If the instruction requires orientation-dependent reasoning, the LLM receives the same information plus explicit orientation angles and their derived directional axes. The exact prompt is:

\begin{tcolorbox}[colback=gray!5,colframe=gray!40,sharp corners]
Text: \texttt{<user\_instruction>}

The main object is \texttt{<main>}.

It has the following candidate positions with indices (x, y, z where z is height):

Index 0: (x=\texttt{<x0>}, y=\texttt{<y0>}, z=\texttt{<z0>})

Index 1: (x=\texttt{<x1>}, y=\texttt{<y1>}, z=\texttt{<z1>})

\ldots

Other objects and their positions are:

\texttt{<obj>}[\texttt{<k>}]: (x=\texttt{<x>}, y=\texttt{<y>}, z=\texttt{<z>})

\ldots
\\

Orientation information (object, label) and angles in degrees:

\texttt{<obj>} (\texttt{<lab>}): [\texttt{<ang>}]
When facing \texttt{<lab>} (\texttt{<ang>}°):
Forward = \texttt{<ang>}° (\texttt{<dirF>}),
Left = \texttt{<left>}° (\texttt{<dirL>}),
Right = \texttt{<right>}° (\texttt{<dirR>}),
Back = \texttt{<back>}° (\texttt{<dirB>}).

Instruction:
Based on the text above, decide which candidate index of \texttt{<main>} is the correct target.
Return only the numeric index (0, 1, 2, \ldots). Do not return anything else.
\end{tcolorbox}

Here, coordinate placeholders (e.g., \texttt{<x0>}, \texttt{<x>}, \texttt{<y>}, \texttt{<z>}) are filled with 3D centroids, name placeholders (e.g., \texttt{<main>}, \texttt{<obj>}) are filled with object identifiers, and orientation placeholders (e.g., \texttt{<lab>}, \texttt{<ang>}, \texttt{<dirF>}, \texttt{<left>}, \texttt{<right>}, \texttt{<back>}) are filled with the yaw angle and its derived directional axes.

\medskip
In both cases, the LLM selects the final target purely from this structured geometric information and outputs a single integer index.

\section{Mask Merging}
We use weighted averaging for our mask merging part. The weight of each mask is determined by its 3D relative position from the camera, amount of semantic the mask contains and the number of pixels in the mask. The position weight is determined by this formula:

\[
w_{d} = 
\begin{cases}
  e^{-\alpha(1.5-d_{avg})} & \text{if } d_{avg} \le 1.5, \\
  1.0 & \text{if } 2.5 \ge d_{avg}  \ge 1.5, \\
  e^{-\alpha(d_{avg}-2.5)}   & \text{if }  d_{avg} \ge 2.5
\end{cases}
\]
The $d_{avg}$ is the average depth of pixels in the mask and $\alpha = 0.5 $.\newline
To measure semantics of each mask, we measure the cosine similarity of each mask with a set of indoor classes texts. The set contains 200 indoor classes and is generated by an LLM. The semantic weight is defined as
$ w_{s} = \displaystyle \max_{s \in S} s $.

The $w_p$ is the number of pixels in the mask. The final weight is calculated by: 
\[
w = w_d \times w_s \times w_p
\]

\section{Hyperparameter Sensitivity}
\label{appendix:hypersensitivity}

\paragraph{Context-aware embedding weights.}
For each 2D object mask we construct five CLIP image embeddings:
\[
E_s,\; E_l,\; E_h,\; E_{\text{hide}},\; E_{\text{mask}} \in \mathbb{R}^d,
\]
corresponding respectively to
(i) a small crop tightly around the object,
(ii) a larger crop including more local context,
(iii) a ``huge'' crop covering the object and its wider surround,
(iv) a crop where the object is removed and only its surroundings are visible,
and (v) an object-only crop obtained by masking out the background.
Given these base embeddings, our normalized context-aware representation for an object is
\begin{equation*}
\label{eq:ctx_embedding}
E_{\text{ctx}}
  = E_s
  + \alpha_h\, E_h
  + \alpha_l\, E_l
  - \alpha_o\, E_{\text{hide}}
  + \alpha_m\, E_{\text{mask}},
\end{equation*}
where the scalar coefficients
$\alpha_h, \alpha_l, \alpha_o, \alpha_m \ge 0$
control how much we emphasize wide-field context ($E_h$),
medium-scale context ($E_l$), suppress misleading background-only evidence
($E_{\text{hide}}$), and sharpen the signal from the object-only view
($E_{\text{mask}}$).

\paragraph{Sensitivity setup.}
In the following experiments we study the effect of these weights on
3D semantic segmentation performance.
For each parameter $\alpha \in \{\alpha_h,\alpha_l,\alpha_o,\alpha_m\}$,
we scale it by a factor $\lambda \in \{0.5, 0.75, 1.0, 1.25, 1.5\}$ while
keeping the remaining coefficients fixed at their default values, and measure mIoU, mAcc, and fmIoU.

\begin{figure*}[h]
  \centering
  \begin{tikzpicture}
    \begin{groupplot}[
      group style={
        group size=3 by 1,
        horizontal sep=2cm,
        vertical sep=1.4cm
      },
      width=0.32\textwidth,
      height=0.24\textwidth,
      xlabel={Scaling factor},
      xmin=0.5, xmax=1.5,
      xtick={0.5,0.75,1.0,1.25,1.5},
      xticklabels={0.5,0.75,1.0,1.25,1.5},
      grid=both,
      grid style={dotted},
      legend style={
        font=\footnotesize,
        at={(0.5,1.15)},
        anchor=south,
        legend columns=4
      },
      legend cell align={left},
      tick label style={font=\footnotesize},
      label style={font=\footnotesize},
      title style={font=\footnotesize}
    ]

    \nextgroupplot[
      ylabel={mIoU}
    ]
      \addplot+[mark=o] coordinates {
        (0.50,0.3094) (0.75,0.2939)
        (1.00,0.3224) (1.25,0.3208) (1.50,0.3207)
      };
      \addlegendentry{$\alpha_h$};

      \addplot+[mark=triangle] coordinates {
        (0.50,0.2833) (0.75,0.2941)
        (1.00,0.3224) (1.25,0.3223) (1.50,0.3201)
      };
      \addlegendentry{$\alpha_l$};

      \addplot+[mark=square] coordinates {
        (0.50,0.3207) (0.75,0.3224)
        (1.00,0.3224) (1.25,0.2939) (1.50,0.3056)
      };
      \addlegendentry{$\alpha_o$};

      \addplot+[mark=diamond] coordinates {
        (0.50,0.2693) (0.75,0.3218)
        (1.00,0.3224) (1.25,0.3396) (1.50,0.3294)
      };
      \addlegendentry{$\alpha_m$};

    \nextgroupplot[
      ylabel={mAcc}
    ]
      \addplot+[mark=o] coordinates {
        (0.50,0.4255) (0.75,0.4038)
        (1.00,0.4330) (1.25,0.4331) (1.50,0.4331)
      };

      \addplot+[mark=triangle] coordinates {
        (0.50,0.3914) (0.75,0.4038)
        (1.00,0.4330) (1.25,0.4352) (1.50,0.4357)
      };

      \addplot+[mark=square] coordinates {
        (0.50,0.4358) (0.75,0.4352)
        (1.00,0.4330) (1.25,0.4038) (1.50,0.4215)
      };

      \addplot+[mark=diamond] coordinates {
        (0.50,0.3903) (0.75,0.4353)
        (1.00,0.4330) (1.25,0.4423) (1.50,0.4304)
      };

    \nextgroupplot[
      ylabel={fmIoU}
    ]
      \addplot+[mark=o] coordinates {
        (0.50,0.5505) (0.75,0.5755)
        (1.00,0.5916) (1.25,0.5909) (1.50,0.5908)
      };

      \addplot+[mark=triangle] coordinates {
        (0.50,0.5496) (0.75,0.5756)
        (1.00,0.5916) (1.25,0.5923) (1.50,0.5888)
      };

      \addplot+[mark=square] coordinates {
        (0.50,0.5893) (0.75,0.5923)
        (1.00,0.5916) (1.25,0.5755) (1.50,0.5405)
      };

      \addplot+[mark=diamond] coordinates {
        (0.50,0.4438) (0.75,0.5916)
        (1.00,0.5916) (1.25,0.6756) (1.50,0.6498)
      };

    \end{groupplot}
  \end{tikzpicture}
  \caption{Hyperparameter sensitivity on Replica (scene \texttt{room0}) for the four weighting parameters $\alpha_h, \alpha_l, \alpha_o, \alpha_m$. Performance is reported in mIoU, mAcc, and fmIoU as the scaling factor is varied.}
  \label{fig:hypersensitivity_replica}
\end{figure*}

\begin{figure*}[h]
  \centering
  \begin{tikzpicture}
    \begin{groupplot}[
      group style={
        group size=3 by 1,
        horizontal sep=2cm,
        vertical sep=1.4cm
      },
      width=0.32\textwidth,
      height=0.24\textwidth,
      xlabel={Scaling factor},
      xmin=0.5, xmax=1.5,
      xtick={0.5,0.75,1.0,1.25,1.5},
      xticklabels={0.5,0.75,1.0,1.25,1.5},
      grid=both,
      grid style={dotted},
      legend style={
        font=\footnotesize,
        at={(0.5,1.15)},
        anchor=south,
        legend columns=4
      },
      legend cell align={left},
      tick label style={font=\footnotesize},
      label style={font=\footnotesize},
      title style={font=\footnotesize}
    ]

    \nextgroupplot[
      ylabel={mIoU}
    ]
      \addplot+[mark=o] coordinates {
        (0.00,0.3008) (0.50,0.3065) (0.75,0.3026)
        (1.00,0.2972) (1.25,0.3137) (1.50,0.3129)
      };
      \addlegendentry{$\alpha_h$};

      \addplot+[mark=triangle] coordinates {
        (0.00,0.3096) (0.50,0.3015) (0.75,0.3026)
        (1.00,0.2972) (1.25,0.3166) (1.50,0.3178)
      };
      \addlegendentry{$\alpha_l$};

      \addplot+[mark=square] coordinates {
        (0.00,0.3182) (0.50,0.3141) (0.75,0.3129)
        (1.00,0.2972) (1.25,0.3086) (1.50,0.3120)
      };
      \addlegendentry{$\alpha_o$};

      \addplot+[mark=diamond] coordinates {
        (0.00,0.2804) (0.50,0.3080) (0.75,0.3185)
        (1.00,0.2972) (1.25,0.3009) (1.50,0.3062)
      };
      \addlegendentry{$\alpha_m$};

    \nextgroupplot[
      ylabel={mAcc}
    ]
      \addplot+[mark=o] coordinates {
        (0.00,0.4844) (0.50,0.4933) (0.75,0.4920)
        (1.00,0.4916) (1.25,0.5085) (1.50,0.5108)
      };

      \addplot+[mark=triangle] coordinates {
        (0.00,0.4829) (0.50,0.4796) (0.75,0.4920)
        (1.00,0.4916) (1.25,0.5148) (1.50,0.5170)
      };

      \addplot+[mark=square] coordinates {
        (0.00,0.5175) (0.50,0.5128) (0.75,0.5078)
        (1.00,0.4916) (1.25,0.4941) (1.50,0.4995)
      };

      \addplot+[mark=diamond] coordinates {
        (0.00,0.4804) (0.50,0.4998) (0.75,0.5135)
        (1.00,0.4916) (1.25,0.4937) (1.50,0.4950)
      };

    \nextgroupplot[
      ylabel={fmIoU}
    ]
      \addplot+[mark=o] coordinates {
        (0.00,0.4171) (0.50,0.4207) (0.75,0.4186)
        (1.00,0.4095) (1.25,0.4247) (1.50,0.4222)
      };

      \addplot+[mark=triangle] coordinates {
        (0.00,0.4240) (0.50,0.4187) (0.75,0.4185)
        (1.00,0.4095) (1.25,0.4260) (1.50,0.4324)
      };

      \addplot+[mark=square] coordinates {
        (0.00,0.4363) (0.50,0.4242) (0.75,0.4225)
        (1.00,0.4095) (1.25,0.4222) (1.50,0.4274)
      };

      \addplot+[mark=diamond] coordinates {
        (0.00,0.3499) (0.50,0.4053) (0.75,0.4261)
        (1.00,0.4095) (1.25,0.4156) (1.50,0.4220)
      };

    \end{groupplot}
  \end{tikzpicture}
  \caption{Hyperparameter sensitivity on ScanNet (scene \texttt{scene0011\_00}) for the four weighting parameters $\alpha_h, \alpha_l, \alpha_o, \alpha_m$. Performance is reported in mIoU, mAcc, and fmIoU as the scaling factor is varied.}
  \label{fig:hypersensitivity_scannet}
\end{figure*}

\newpage

\section{Visualization}
We include the following visualizations to enable a more detailed inspection of the results across different Replica scenes, facilitate comparison across methods, and to illustrate typical failure cases observed in challenging scenarios.

\begin{figure}[h]
    \centering
    \makebox[\textwidth]{%
        \includegraphics[width=1\textwidth]{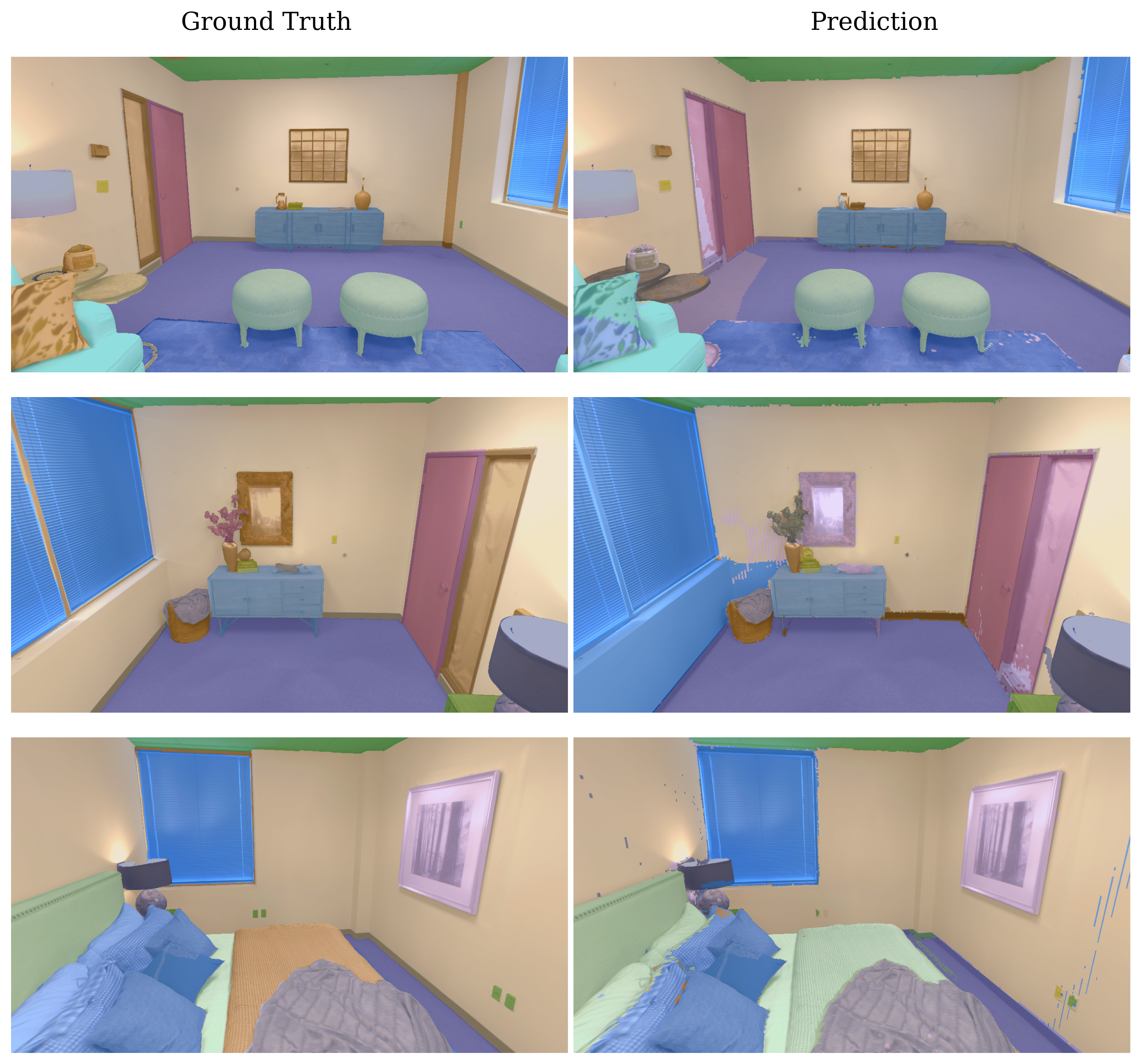}
    }
\end{figure}

\begin{figure}[h]
    \centering
    \makebox[\textwidth]{%
        \includegraphics[width=1\textwidth]{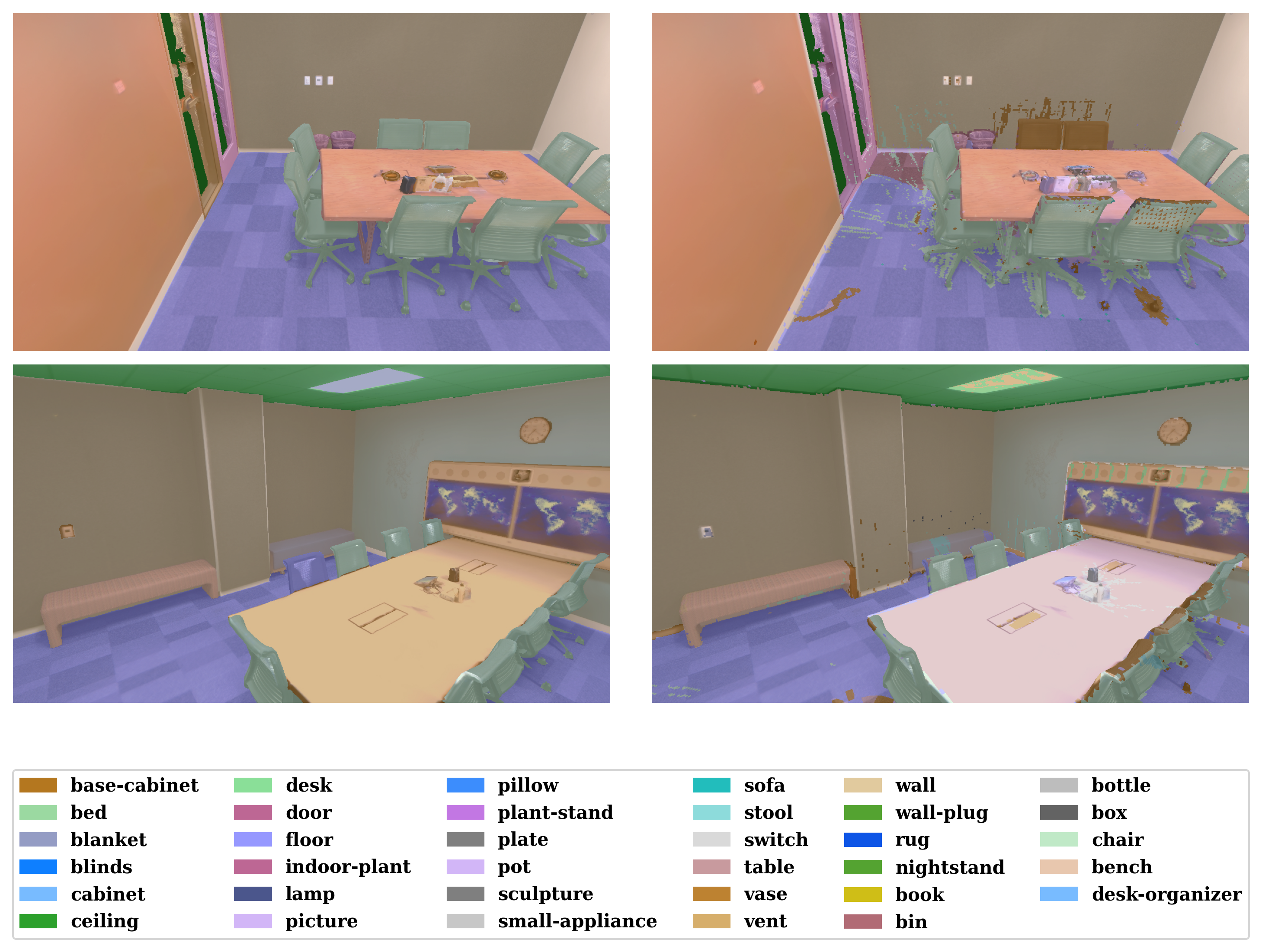}
    }
\end{figure}

\end{document}